\documentclass{article}
\usepackage{spconf, amsmath, graphicx, booktabs, multirow, amsfonts, subcaption, caption}
\usepackage{tabularx}

\title{MHLAT: Multi-hop Label-wise Attention Model for Automatic ICD Coding}
\name{Junwen Duan$^{1,2}$, Han Jiang$^{1}$, Ying Yu$^{1}$}%\thanks{This work is partly supported by the National Natural Science Foundation of China (No. 62006251), the Natural Science Foundation of Hunan Province (No. 2021JJ40783).}}
\address{$^{1}$School of Computer Science and Engineering, Central South University, Changsha, Hunan, China\\
$^{2}$Hunan Provincial Key Lab on Bioinformatics, Central South University, Changsha, Hunan, China\\
\textit{\{jwduan, jh-better, yuying\}@csu.edu.cn}
}

\begin{document}
% \topmargin=0mm
%\ninept
%
\maketitle
\begin{abstract}
International Classification of Diseases (ICD) coding is the task of assigning ICD diagnosis codes to clinical notes. This can be challenging given the large quantity of labels (nearly 9,000) and lengthy texts (up to 8,000 tokens). However, unlike the single-pass reading process in previous works, humans tend to read the text and label definitions again to get more confident answers. Moreover, although pretrained language models have been used to address these problems, they suffer from huge memory usage. To address the above problems, we propose a simple but effective model called the \textit{\textbf{M}ulti-\textbf{H}op \textbf{L}abel-wise \textbf{AT}tention~(MHLAT)}, in which multi-hop label-wise attention is deployed to get more precise and informative representations. Extensive experiments on three benchmark MIMIC datasets indicate that our method achieves significantly better or competitive performance on all seven metrics, with much fewer parameters to optimize.

\end{abstract}
\vspace{-0.5em}
\section{Introduction}
\vspace{-0.5em}
\label{sec:intro}
% International Classification of Diseases (ICD) coding is an international unified disease classification method developed by the World Health Organization, whose code consists of letters and numbers. 
% Generally doctors assign ICD codes to clinical notes to describe diagnoses and procedures that were performed during treatment.
International Classification of Diseases (ICD) coding uses letters and numbers to classify diseases internationally. 
Diagnoses and procedures performed during treatment are described with ICD codes in clinical notes.
Due to the labor-intensive and error-prone classification procedure by humans, many deep learning methods~\cite{HaoranShi2017TowardsAI, JamesMullenbach2018ExplainablePO, ThanhVu2020ALA, LeiboLiu2022HierarchicalLA} have been applied to automate ICD code assignment.

Suffering from extremely long input texts and highly large label spaces with long-tail label distribution, instead of using pre-trained language models(PLMs)~\cite{JacobDevlin2018BERTPO}, many researchers allocate typical CNNs~\cite{JamesMullenbach2018ExplainablePO,  FeiLi2019ICDCF,YangLiu2021EffectiveCA} or RNNs~\cite{HaoranShi2017TowardsAI, ThanhVu2020ALA} to model the n-grams relationship (CNNs) or the sequential dependency (RNNs).
A recent CNN approach is Effective CAN~\cite{YangLiu2021EffectiveCA} which borrows the squeeze and excitation mechanism from SENet~\cite{JieHu2018SqueezeandExcitationN} and combines features with different receptive fields to produce the final codes.
LAAT~\cite{ThanhVu2020ALA} is a state-of-the-art method based on RNN which uses bidirectional long short-term memory (BiLSTM)~\cite{SeppHochreiter1997LongSM} to directly model sequential dependency from original texts.
Because of the lack of powerful feature extractors, e.g. BERT~\cite{JacobDevlin2018BERTPO}, the model performance is thus limited.
% BERT-ICD~\cite{DamianPascual2021TowardsBA} argues BERTs~\cite{JacobDevlin2018BERTPO} are not suitable for ICD coding because of the extremely long input sequences.
HiLAT~\cite{LeiboLiu2022HierarchicalLA} is the first work to successfully use PLMs into ICD coding and achieves the best results on MIMIC-III 50 dataset~\cite{AlistairEWJohnson2016MIMICIIIAF}.
However, HiLAT~\cite{LeiboLiu2022HierarchicalLA} fails to train on MIMIC-III full dataset~\cite{AlistairEWJohnson2016MIMICIIIAF} which has large label spaces, posing a challenge to device memory. 
Additionally, previous methods used single-pass attention, which is counterintuitive, as humans tend to first read the text roughly and then again after confirming the label information to obtain more accurate and informative representations.

% To overcome above problems, we propose a simple yet effective approach named Multi-hop Label-wise Attention model.
% In particular, our model a) deploy a multi-hop label-wise attention to mimic the human reading process, and b) introduce Bias-Terms Fine-Tuning~(BitFit)~\cite{ben-zaken-etal-2022-bitfit} in which only the bias terms are fine-tuned to reduce the parameters to tune.
% We summarize the contributions of our work as:
To overcome the above problems, we propose a simple yet effective approach named the Multi-hop Label-wise Attention model. Specifically, our model a) uses a multi-hop label-wise attention mechanism to mimic the human reading process, and b) introduces Bias-Terms Fine-Tuning (BitFit)~\cite{ben-zaken-etal-2022-bitfit}, in which only the bias terms are fine-tuned to reduce the number of parameters that need to be tuned.

We summarize the contributions of our work as follows:
\begin{itemize}
\vspace{-0.5em}
\item We propose multi-hop label-wise attention model to mimic the human reading process.%deal with inaccurate information extraction problem.

\vspace{-0.5em}
\item We deploy Bias-Terms Fine-Tuning~(BitFit) to reduce the number of parameters in PLMs to tune.

\vspace{-0.5em}
\item We evaluate our method on three MIMIC datasets commonly used for ICD coding. Extensive experiments indicate that our method is effective and achieves state-of-the-art performance.

\end{itemize}

\begin{table}[]
\small
\centering
\begin{tabular}{c|ccc}
\toprule
\multirow{2}{*}{}     & \multicolumn{2}{c|}{MIMIC-III}                      & MIMIC-II \\ \cline{2-4} 
                      & \multicolumn{1}{c|}{full} & \multicolumn{1}{c|}{50} & full     \\ \hline
avg \#tokens/doc & 1483                      & 1527                    & 1106     \\
max \#tokens/doc & 8772                      & 7567                    & 6231     \\
\#labels           & 8922                      & 50                      & 5031    \\ \toprule
\end{tabular}
\caption{Statistics of all three benchmark MIMIC datasets.}
\label{dataset}
\vspace{-1em}
\end{table}

\vspace{-1.5em}
\section{Problem Definition}
% \vspace{-0.5em}
We formulate the ICD coding assignment problem as a multi-label classification task. 
\begin{figure*}
    \vspace{-1em}
    \centering
    \includegraphics[width=6 in]{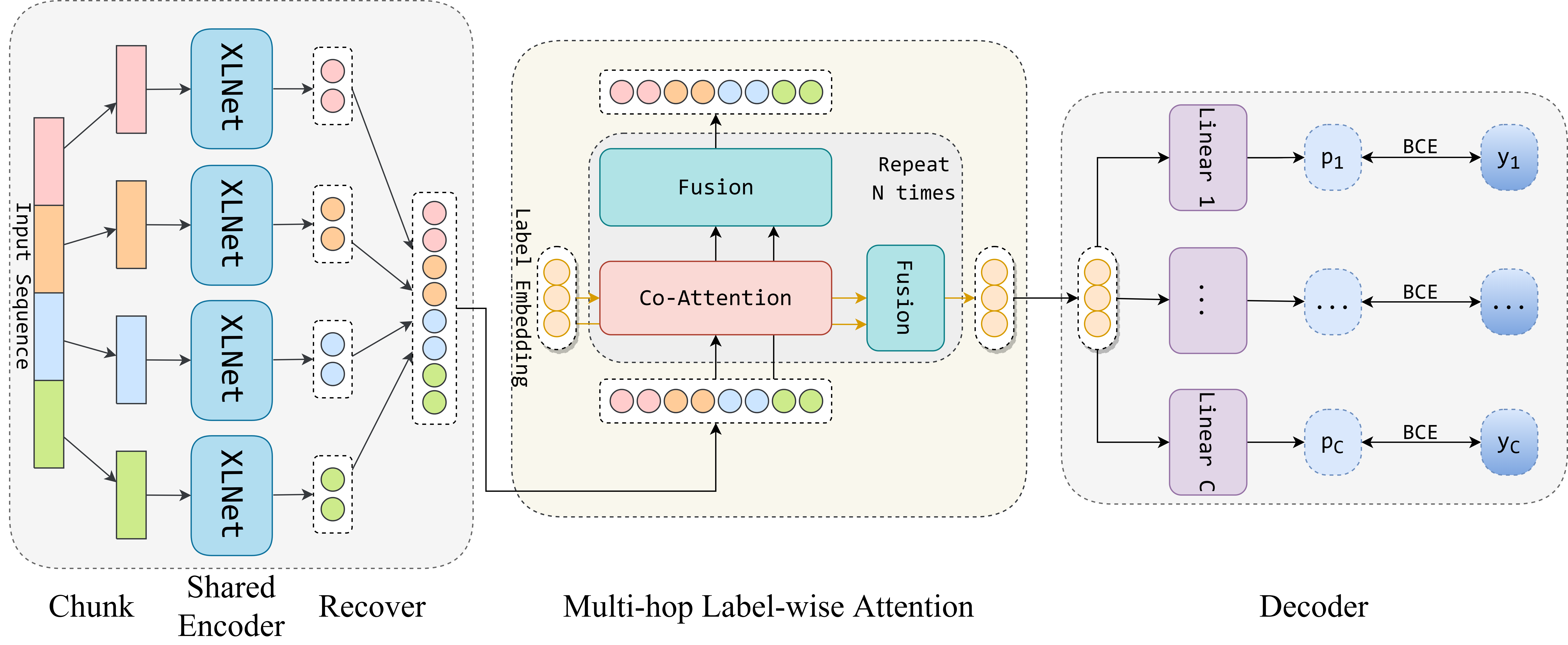}
    \caption{
    The framework of our approach. The clinical notes are chunked into $k$ ($k=4$ in the example) segments and fed to the shared PLM encoder. The features are then recovered for global representation. Multi-hop label-wise attention is performed to obtain label representations. Finally, label-specific classifier is deployed for label scoring.
    % Our model architecture. Firstly, a long sentence are chunked into k (k=4 for example) segments and input to PLM seperately. Secondly, we concatenate the features of all k segments sequentially and pass recovered features into label-wise attention for label expression. Finally, a individual linear layer then applied to get final prediction.
    }
    \vspace{-1em}
    \label{model}
\end{figure*}
Given an input sequence \(\mathbf{x}=\{x_1, x_2, ..., x_n\}\)
% whose corresponding token representation is \(\mathbf{W}=\{\mathbf{w_1}, \mathbf{w_2}, ..., \mathbf{w_n}\} \in \mathbb{R} ^ {d_e\times n}\) where $ d_e $ denotes the token embedding dimensions.
, we aim to classify it into corresponding labels \(\mathbf{y} \in \{0,1\}^C\), where $n$ indicates the number of input tokens and $C$ denotes the size of the entire label space, e.g. 50 in the MIMIC-III 50 dataset.
The text and label pair \((\mathbf{x}, \mathbf{y})\) form a sample in the dataset $\mathcal{D}$.

\begin{table}[]
\small
\centering
\begin{tabular}{l|llll}
\toprule
Models                                                          & Macro F1 & Micro F1 & P@8  & P@15 \\ \hline
CAML\cite{JamesMullenbach2018ExplainablePO}    & 4.8      & 44.2     & 52.3 & -    \\
DR-CAML\cite{JamesMullenbach2018ExplainablePO} & 4.9      & 45.7     & 51.5 & -    \\
MultiResCNN\cite{FeiLi2019ICDCF}               & 5.2      & 46.4     & 54.4 & -    \\
HyperCore\cite{PengfeiCao2020HyperCoreHA}      & 7.0      & 47.7     & 53.7 & -    \\
LAAT\cite{ThanhVu2020ALA}                      & 5.9      & 48.6     & 55.0 & 39.7 \\
JointLAAT\cite{ThanhVu2020ALA}                 & 6.8      & 49.1     & 55.1 & 39.6 \\
ISD\cite{TongZhou2021AutomaticIC}              & \textbf{10.1}     & 49.8     & 56.4 & -    \\ \hline
MHLAT(\textit{ours})                           & 8.9      & \textbf{51.0}     & \textbf{57.1} & \textbf{41.5} \\ \toprule
\end{tabular}
\caption{Results (in \%) on MIMIC-II full test set. \textbf{Bold} indicates the best result among baselines. }
\label{table-2full} 
\vspace{-1.5em}
\end{table}

\vspace{-1em}
\section{Method}
\vspace{-0.5em}

The overall architecture of \textit{\textbf{M}ulti-\textbf{H}op \textbf{L}abel-wise \textbf{AT}tention (MHLAT)} model is shown in Figure~\ref{model}, which comprises four main components: a chunk and input layer, an encoding layer, a multi-hop label-wise attention layer, and a decoding layer. 
% The long input sequence is first chunked into $k$ chunks in chunk and input layer and fed into encoder layer to get semantic features.
The long input sequence is first chunked into $k$ chunks in the chunk and input layer and then fed into the encoder layer to obtain semantic features. 
% Powerful label features are then extracted by multi-hop label-wise attention layer and used to compute final prediction scores in decoding layer.
% The details of each component are introduced in the following sections. 
Subsequently, powerful label features are extracted by the multi-hop label-wise attention layer. 
Finally, these label features are used to compute the final prediction scores in the decoding layer. 
In the following sections, we will introduce the details of each component.

\vspace{-0.5em}
\subsection{Chunk and Input Layer}
Traditional PLMs have difficulty handling long text since both the space and time complexities are quadratic.%~\cite{AshishVaswani2017AttentionIA}
Although some works are proposed to address the problem, there are currently no in-domain PLMs in the medical domain. We propose the Chunk and Input layer to handle this problem instead.

For an extremely long sequence \(\mathbf{x}=\{x_1,x_2,...,x_n\}\), we limit the maximum length $L$ for each chunk, e.g. $L=512$, then we obtain $k$ chunks with \(\{\mathbf{x}^{(1)}, \mathbf{x}^{(2)},...,\mathbf{x}^{(k)}\}\) and their corresponding token embeddings \(\{\mathbf{W}^{(1)}, \mathbf{W}^{(2)},...,\mathbf{W}^{(k)}\}\), where \(\mathbf{W}^{(i)}\in \mathbb{R}^{L\times d_{w}}\). Here, \(d_w\) denotes the dimension of the token embedding.

\vspace{-0.5em}
\subsection{Encoding Layer}
Previous methods mainly focused on traditional deep learning encoders, such as CNNs and RNNs.
However, such encoders fail to capture long range dependency and have limited ability to extract information from texts.
Pre-trained language models have been proposed to address these problems.
% Furthermore, to diminish gaps between general domain and specific domain, researchers often further pre-train the PLMs on the specific domain corpus, such as medical domain.
Moreover, to bridge the gap between general domain and specific domain, researchers often perform further pre-training of PLMs on specific domain corpora, such as the medical domain.

Therefore, we adopt the pre-trained XLNet~\cite{ZhilinYang2019XLNetGA} from~\cite{LeiboLiu2022HierarchicalLA} as the encoder, which is further pre-trained on the MIMIC corpus. 
The encoder is parameter-shared and apply to each chunk, as illustrated in Equation~(\ref{eq:encoder}).
\begin{align}
    \mathbf{h}^{(i)}&=\text{xlnet}(\mathbf{W}^{(i)})\in\mathbb{R}^{L\times d_m}\text{, for $i$ in }1,2,3,...,k.
    \label{eq:encoder}
\end{align}
where $d_m$ is the output dimension of XLNet.
It's noteworthy that the BitFit algorithm~\cite{ben-zaken-etal-2022-bitfit} is deployed on XLNet backbone for parameter-efficient training.

After encoding each chunk, we obtain chunk features \(\{\mathbf{h}^{(1)},\mathbf{h}^{(2)},...,\mathbf{h}^{(k)}\}\).
We then concatenate them sequentially to get the global information of the input sentence $\mathbf{x}$.
\begin{align}
    \mathbf{H}&=concat(\mathbf{h}^{(1)}, \mathbf{h}^{(2)}, ..., \mathbf{h}^{(k)}) \in \mathbb{R}^{n\times d_m}.
\end{align}

\begin{table}[]
\small
\vspace{-1em}
\centering
\begin{tabular}{l|llll}
\toprule
Models                                                          & Macro F1 & Micro F1 & P@8  & P@15 \\ \hline
CAML\cite{JamesMullenbach2018ExplainablePO}    & 8.8      & 53.9     & 70.9 & 56.1 \\
DR-CAML\cite{JamesMullenbach2018ExplainablePO} & 8.6      & 52.9     & 69.0 & 54.8 \\
MultiResCNN\cite{FeiLi2019ICDCF}               & 8.5      & 55.2     & 73.4 & 58.4 \\
MSATT-KG\cite{XianchengXie2019EHRCW}           & 9.0      & 55.3     & 72.8 & 58.1 \\
HyperCore\cite{PengfeiCao2020HyperCoreHA}      & 9.0      & 55.1     & 72.2 & 57.9 \\
EffectiveCAN\cite{YangLiu2021EffectiveCA}      & 10.6     & 58.9     & 75.8 & 60.6 \\
LAAT\cite{ThanhVu2020ALA}                      & 9.9      & 57.5     & 73.8 & 59.1 \\
JointLAAT\cite{ThanhVu2020ALA}                 & 10.7     & 57.5     & 73.5 & 59.0 \\
MARN\cite{WeiSun2021MultitaskBA}              & 11.6     & 58.4     & 75.4 & 60.2 \\
RAC\cite{ByungHakKim2021ReadAA}               & \textbf{12.7}     & 58.6     & 75.4 & 60.1 \\
% MSMN\cite{ZhengYuan2022CodeSD}                & 10.3     & 58.4     & 75.2 & 59.9 \\
MDBERT\cite{NingZhang2022HierarchicalBF}      & 10.4     & 57.6     & 75.0 & 59.6 \\
ISD\cite{TongZhou2021AutomaticIC}              & 11.9     & 55.9     & 74.5 & -    \\ \hline
MHLAT(\textit{ours})                           & 11.2     & \textbf{59.1}     & \textbf{75.9} & \textbf{61.3} \\ \toprule
\end{tabular}
\caption{Results (in \%) on MIMIC-III full test set. \textbf{Bold} indicates the best result among baselines. }
\label{table-3full}
\vspace{-1.5em}
\end{table}

\vspace{-0.5em}
\subsection{Multi-hop Label-wise Attention Layer}
While the traditional attention mechanism performs poorly on multi-label classification problems, LAAT~\cite{ThanhVu2020ALA} proposed a label-wise attention mechanism to extract individual information from contextualized features for each label, which provides a big performance boost. 
However, single-pass classification is error-prone and humans tend to reread passages and label definitions to ensure the correct labels are assigned to the text. 
To address this, we propose a multi-hop label-wise attention mechanism to refine label representations and capture more informative features. %and second  the label-specific representation. 

% \begin{figure}
%     \centering
%     \includegraphics[width=3 in]{attention.png}
%     \caption{
%     Our multi-hop label-wise attention mechanism. 
%     }
%     \label{attention}
% \end{figure}

% We propose another form of label-wise attention here to further enhance the label representations.

\begin{table*}[htb]
\small
\vspace{-1em}
\centering
\begin{tabular}{ll|ll|ll|lll}
\toprule
\multicolumn{2}{l|}{\multirow{2}{*}{Methods}}                                                   & \multicolumn{2}{c|}{AUC}      & \multicolumn{2}{c|}{F1}       & \multicolumn{3}{c}{P@k}                       \\ \cline{3-9} 
\multicolumn{2}{l|}{}                             & Macro         & Micro         & Macro         & Micro         & P@5           & P@8           & P@15          \\ \hline
\multicolumn{1}{l|}{}         & CAML~\cite{JamesMullenbach2018ExplainablePO}                                       & 87.5          & 90.9          & 53.2          & 61.4          & 60.9          & -             & -             \\
\multicolumn{1}{l|}{}                             & DR-CAML~\cite{JamesMullenbach2018ExplainablePO}                                     & 88.4          & 91.6          & 57.6          & 63.3          & 61.8          & -             & -             \\
\multicolumn{1}{l|}{}                             & MultiResCNN~\cite{FeiLi2019ICDCF}                                 & 89.9          & 92.8          & 60.6          & 67.0          & 64.1          & -             & -             \\
\multicolumn{1}{l|}{}                             & MSATT-KG~\cite{XianchengXie2019EHRCW}                                    & 91.4          & 93.6          & 63.8          & 68.4          & 64.4          & -             & -             \\
\multicolumn{1}{l|}{}                             & HyperCore~\cite{PengfeiCao2020HyperCoreHA}                                   & 89.5          & 92.9          & 60.9          & 66.3          & 63.2          & -             & -             \\
\multicolumn{1}{l|}{}                             & EffectiveCAN~\cite{YangLiu2021EffectiveCA}                                & 92.0          & 94.5          & 66.8          & 71.7          & 66.4          & -             & -             \\ 
% \multicolumn{1}{l|}{\multirow{3}{*}{RNN}}         & C-LSTM-Att~\cite{HaoranShi2017TowardsAI}                                  & -             & 90.0          & -             & 53.2          & -             & -             & -             \\
\multicolumn{1}{l|}{\multirow{2.3}{*}{Baselines}}                             & LAAT~\cite{ThanhVu2020ALA}                                        & 92.5          & 94.6          & 66.6          & 71.5          & 67.5          & 54.7          & 35.7          \\
\multicolumn{1}{l|}{}                             & JointLAAT~\cite{ThanhVu2020ALA}                                   & 92.5          & 94.6          & 66.1          & 71.6          & 67.1          & 54.6          & 35.7          \\
% \multicolumn{1}{l|}{} & BERT-ICD~\cite{DamianPascual2021TowardsBA}                                    & 84.5          & 88.7          & -             & -             & -             & -             & -             \\
\multicolumn{1}{l|}{}                             & Longformer-DLAC~\cite{MalteFeucht2021DescriptionbasedLA}                             & 87.0          & 91.0          & 52.0          & 62.0          & 61.0          & -             & -             \\
\multicolumn{1}{l|}{}                             & MSMN~\cite{ZhengYuan2022CodeSD}                                    & 92.8          & 94.7          & 68.3          & 72.5          & 68.0          & -             & -             \\
\multicolumn{1}{l|}{}                             & MARN~\cite{WeiSun2021MultitaskBA}                                    & 92.7          & 94.7          & 68.2          & 71.8          & 67.3          & -             & -             \\
\multicolumn{1}{l|}{}                             & MDBERT~\cite{NingZhang2022HierarchicalBF}                                    & 92.8          & 94.6          & 67.2          & 71.7          & 67.4          & -             & -             \\
\multicolumn{1}{l|}{}                             & ISD~\cite{TongZhou2021AutomaticIC}                                         & \textbf{93.5} & 94.9          & 67.9         & 71.7          & 68.2          & -             & -             \\
\multicolumn{1}{l|}{}                             & HiLAT~\cite{LeiboLiu2022HierarchicalLA}                                       & 92.7          & 95.0         & 69.0 & 73.5         & 68.1         & 55.4         & 36.2         \\ \hline 
\multicolumn{1}{l|}{\multirow{8}{*}{Ablation Studies}}                             & MHLAT(\textit{ours})                                        & 93.1         & \textbf{95.1} & \textbf{69.2}          & \textbf{73.9} & \textbf{68.7} & \textbf{55.8} & \textbf{36.3} \\
\multicolumn{1}{l|}{}                             & \textit{-2 hop}                                      & 72.3         & 72.6 & 7.8          & 18.0 & 34.6 & 29.0 & 23.2 \\
\multicolumn{1}{l|}{}                             & \textit{-1 hop}                                        & 93.1         & \textbf{95.1} & 68.5          & 73.7 & 68.3 & \textbf{55.8} & 36.2 \\
\multicolumn{1}{l|}{}                             & \textit{+1 hop}                                      & 92.9         & \textbf{95.1} & 69.1          & 73.3 & 68.3 & 55.7 & \textbf{36.3} \\
\multicolumn{1}{l|}{}                             & \textit{+2 hop}                                      & \textbf{93.1}         & \textbf{95.1} & 68.2          & 73.2 & 68.4 & 55.7 & \textbf{36.3} \\
\multicolumn{1}{l|}{}                             & \textit{+share params}                                      & 92.0         & 94.1 & 64.8          & 70.9 & 66.0 & 54.0 & 35.6 \\
\multicolumn{1}{l|}{}                             & \textit{-BitFit(freeze)}                                      & 91.9         & 94.3 & 64.0          & 70.1 & 66.4 & 53.9 & 35.7 \\
\multicolumn{1}{l|}{}                             & \textit{-BitFit(fine-tune)}                                      & 92.6         & 95.0 & 68.0          & 73.1 & 68.0 & 55.3 & 36.1 \\
\toprule
\end{tabular}
\caption{\label{table-50} Results (in \%) on MIMIC-III 50 test set. \textbf{Bold} indicates the best result among baselines. }
\vspace{-1.5em}
\end{table*}

For the hidden features $\mathbf{H}$ output from the encoder and randomly initialized label embeddings $\mathbf{E}\in \mathbb{R}^{C\times d_{m}}$, we derive our label-wise attention as follows:
% \begin{center}
\begin{gather}
    \mathbf{T}=\text{relu}(\mathbf{W}^{att}\mathbf{H})\cdot \text{relu}(\mathbf{W}^{att}\mathbf{E})^\top \label{func_start}\\
    \alpha=\text{softmax}(\mathbf{T})\\
    \mathbf{Z}=\alpha \cdot \mathbf{H}\in \mathbb{R}^{C\times d_m}
\end{gather}
% \end{center}
where \(\mathbf{W}^{att}\in \mathbb{R}^{d_m\times d_m}\) are trainable parameters, \(\mathbf{Z} \in \mathbb{R}^{C\times d_m}\) represent label-specific representations with \(\mathbf{z}_i\) indicating $i$th label vector conditioned on input text and $C$ denotes the number of labels. 

In order to learn much meaningful label embedding, we then employ the fusion operation which combines the label-specific representation and the label embedding and maps them by passing them through a linear layer.
\begin{gather}
    \mathbf{\tilde{E}}=\mathbf{W}^{map}[\mathbf{Z};\mathbf{E}]+\mathbf{b}^{map}\in\mathbb{R}^{C\times d_m} \label{fusion}
\end{gather}
Where $\mathbf{W}^{map}\in \mathbb{R}^{d_m\times 2d_m}$ and $\mathbf{b}^{map}\in \mathbb{R}^{d_m}$ are trainable parameters of the fusion operation.

When we read the context for the first time, we don't have any label information in our mind, which results in focusing on the wrong places.
To solve this problem, we update the context vector $\mathbf{H}$ conditioned on label embedding $\mathbf{E}$. 
Additionally, the same linear fusion operation described in Equation~(\ref{fusion}) is applied on $\mathbf{D}$ and $\mathbf{H}$:
\begin{gather}
    \beta=\text{softmax}(\mathbf{T}^\top) \\
    \mathbf{D}=\beta \cdot \mathbf{E} \in\mathbb{R}^{n\times d_m}\\
    \mathbf{\tilde{H}}=\mathbf{W}^{map}[\mathbf{D};\mathbf{H}]+\mathbf{b}^{map} \in\mathbb{R}^{n\times d_m}\label{func_end}
\end{gather}

We denote Equation~(\ref{func_start})-(\ref{func_end}) as a hop function, which takes the context information $\mathbf{H}$ and label embedding $\mathbf{E}$ as inputs and outputs more informative context information $\mathbf{\tilde{H}}$ and label embedding $\mathbf{\tilde{E}}$, i.e. $\mathbf{\tilde{H}}, \mathbf{\tilde{E}}=\text{HOP}(\mathbf{H}, \mathbf{E})$. 

Intuitively, we will repeat the $\text{HOP}$ function again to perform the two-hops attention to imitate the human reading process. 
However, our $\text{HOP}$ function remains the shape of inputs and outputs unchanged.
Thus, theoretically, we can perform multi-hop operation to get more powerful contextual information.
In the multi-hop setting, given $\mathbf{H}^{[l]}$ and $\mathbf{E}^{[l]}$ representing the outputs from the $l$-th hop, we can compute the outputs of $(l+1)$-th hop using following formula:
\begin{gather}
    \mathbf{H}^{[l+1]}, \mathbf{E}^{[l+1]}=\text{HOP}^{[l+1]}(\mathbf{H}^{[l]},\mathbf{E}^{[l]})
\end{gather}
We set $\mathbf{H}^{[0]}=\mathbf{H}$ and $\mathbf{E}^{[0]}=\mathbf{E}$ as the inputs for the first hop.
We can also share the same parameters among different hops, which could reduce the number of parameters while increasing the optimization difficulty.
In our full datasets, the parameters are shared to prevent overfitting, while we do not do so in the MIMIC-50 dataset.

Finally, if we deploy N-hops attention, we'll treat the label representation of the final hop $\mathbf{E}^{[N]}$ as the output of multi-hop label-wise attention layer.
By default, two-hops attention is deployed in our model.

\vspace{-0.5em}
\subsection{Decoding Layer}
After get the informative label representations, we use label-specific classifiers to determine whether the input text belongs to a label or not.
For each classifier, we use an independent linear layer for computing the label score.
% Finally, binary cross-entropy loss can be optimized by gradient descend algorithm.
Finally, the binary cross-entropy loss can be formulated as follows:
\begin{gather}
    \hat{y}_i=\mathbf{w}^{cls}_i\mathbf{e}^{[N]}_i+b^{cls}_i\in\mathbb{R}\text{, for $i$ in }1,2,3,...,C\\
    % L(\hat{y}_i, y_i)=-y_i\cdot\log(\hat{y}_i)&-(1-y_i)\cdot\log(1-\hat{y}_i),\\
    % &\text{for $i$ in }1,2,3,...,m.\\
    % \text{linear}(\mathbf{z}_i),\\
    L(\hat{y}_i, y_i)=-\sum_{i=1}^{C}(y_i\log\hat{y}_i+(1-y_i)\log(1-\hat{y}_i)) \label{decoder}
\end{gather}
where \(\mathbf{W}^{cls}=(\mathbf{w}^{cls}_1, \mathbf{w}^{cls}_2, \cdots, \mathbf{w}^{cls}_C)\in\mathbb{R}^{C\times d_m}\) and \(\mathbf{b}^{cls}=(b^{cls}_1, b^{cls}_2, \cdots, b^{cls}_C)\in\mathbb{R}^C\) are trainable parameters of label-specific classifiers.

\vspace{-1em}
\section{Experiments and Results}
\vspace{-0.5em}
\subsection{Datasets and Metrics}
We evaluate our approach on the benchmark MIMIC-II~\cite{JoonLee2011OpenaccessMD} and  MIMIC-III~\cite{AlistairEWJohnson2016MIMICIIIAF}, which are open datasets of ICU medical records.
As shown in Table~\ref{dataset}, the length of input sequences is quite long (up to 8,000 tokens) and the label space size is relatively large (nearly 9,000 labels).
% from \textbf{M}edical \textbf{I}nformation \textbf{M}art for \textbf{I}ntensive \textbf{C}are (MIMIC) benchmark
% which are the difficulties of using PLMs.

We use the same train-val-test partition and evaluation metrics as~\cite{ThanhVu2020ALA} for fair comparisons. 
The metrics includes macro- and micro-F1, macro- and micro-AUC (Area Under the ROC Curve) and Precision@k (\(k \in \{5,8,15\}\)).
% More statistical results for all three datasets are available in Appendix~\ref{sec:appendix}.

% \subsection{Hyperparameters}
% We train our model with AdamW~\cite{IlyaLoshchilov2017DecoupledWD} and set its learning rate to 7e-4 and weight decay to 0.1.
% The batch size is set to 16 and the mini-batch size is set to 2. 
% We totally train our model 10 epoch with linear warm-up and decay deployed.
% The number of warm-up steps is set to 505 which is exactly the number of one epoch training steps on the MIMIC-III 50 training set.
% We also deployed dropout with a drop rate of 0.1 after encoder and label-wise attention.
% we keep our hyperparameters permanent across all datasets which were searched from MIMIC-III 50 dataset.
% More parameter selection information can be found in {APPENDIX}.

% To make a comprehensive comparison with previous work, we directly use the same evaluation metrics with~\cite{ThanhVu2020ALA}, which includes macro- and micro-f1, macro- and micro-AUC (Area Under the ROC Curve) and Precision@k (\(k \in \{5,8,15\}\)).
\vspace{-0.5em}
\subsection{Results}

% \begin{figure*}[htb]
%     \centering
%     \subfloat[\label{chunk}]{
% 		\includegraphics[width=3 in]{chunk.jpg}}
% 	\subfloat[\label{backbone}]{
% 		\includegraphics[width=3 in]{backbone.jpg}}
%     % \includegraphics[width=3 in]{EMNLP 2022/chunk.jpg}
%     \caption{
%     The result on MIMIC-III 50 test set with (a) different maximum chunk lengths and (b) different encoders.
%     % Note that we keep the total number of tokens of each experiment permanent in (a).
%     }
%     % \label{chunk}
%     % \end{subfigure}
% \end{figure*}

Experimental results on MIMIC-III 50 dataset are shown in Table~\ref{table-50}. 
Our method outperforms all other baselines and achieves a new state-of-the-art performance.

Our results on the full datasets are shown in Table~\ref{table-2full} and Table~\ref{table-3full}.
% We observe that transformer-based methods~\cite{TongZhou2021AutomaticIC,LeiboLiu2022HierarchicalLA} are commonly better than RNN/CNN-based methods~\cite{JamesMullenbach2018ExplainablePO,ThanhVu2020ALA}.
% We believe an important reason for the improvement is that PLMs are more powerful encoder than CNNs/RNNs.
% ISD~\cite{TongZhou2021AutomaticIC} uses shared attention and transformer decoder to learn the dynamic code co-occurrence patterns, which is better than HyperCore~\cite{PengfeiCao2020HyperCoreHA} with improvements in macro- and micro-f1 by 3.1\% and 2.2\%, respectively. 
In both full datasets, our method produces higher results in micro-f1, P@8, and P@15 and achieves slightly worse macro-f1 than others, because our model do not explicitly models the long tail distributions of the labels and thus performs worse on rare labels.
We will solve this problem in our future work.

\vspace{-0.5em}
\subsection{Ablation Study}

\textbf{Effectiveness of BitFit}
We also compared the performance of BitFit along with a froze and a fine-tuned model for the verification of necessity of BitFit. 
As shown in Table~\ref{table-50}, both fine-tune and BitFit perform better than froze model, suggesting that fine-tuning the PLM should improve performance on downstream tasks. 
% Interestingly, BitFit achieves better performance than fine-tuning. 
Moreover, our experiments reveal that BitFit achieves even better performance than fine-tuning, demonstrating the effectiveness of our method in adapting pre-trained models to specific downstream tasks while minimizing the amount of training parameters.

\noindent\textbf{The Influence of Hops}
Two-hops attention is deployed in our model by default. 
We also explore the influence of different hops, the results have shown in Table~\ref{table-50}, \textit{-1 hop} and \textit{+1 hop} represent one-hop and three-hops attention, respectively.
Two-hops attention does improve model performance compared to one-hop one, while three-hops achieves the worst performance.
We also observed that three-hops attention model converge slowly in our experiments.
% We believe one important reason for above problems is too many hops will overfit the training data and result in difficulty in optimizing the model.
An important reason for the above problems is that too many hops will result in overfitting and make optimizing the model difficult.

\noindent\textbf{Share Parameters among Hops}
As shown in Table~\ref{table-50}, sharing parameters among hops in the MIMIC-50 dataset results in worse performance because it limits the model's capability.
However, in our full datasets, which have large label spaces and thus introduce a large amount of parameters, sharing parameters works better than using independent ones, which successfully prevent our model from overfitting.

\vspace{-1em}
\section{Conclusion}
% \vspace{-0.5em}
\label{sec:bibtex}
% ICD coding is a labor-intensive and error-prone task and researchers tend to solve is using deep learning.
% Due to the large number of parameters of PLMs and long input sequences the datasets have, PLMs fail to train when the number of labels increases.
% In this paper, we proposed multi-hop label-wise attention model for the ICD coding classification in which our model explicitly mimics human reading process.
% Our experiments on three benchmark MIMIC datasets showed that our method is effective, which yield new state-of-the-art performance on this task. 
In this paper, we have presented a multi-hop label-wise attention model for the task of ICD coding classification.
Our model leverages the hierarchical structure of the label space and explicitly mimics the human reading process to achieve state-of-the-art performance on three benchmark MIMIC datasets. We have also shown the effectiveness of BitFit technique and the impact of different hops of attention mechanism on the model performance. 
% However, our model still faces challenges in dealing with rare labels and long-tail label distributions. 
% Future work can focus on improving the model's capability to handle such cases, as well as exploring the application of our proposed approach to other related tasks in the medical domain.

\vspace{-1em}
\section{acknowledgement}
% \vspace{-0.5em}
This work is partly supported by the National Natural Science Foundation of China (No. 62006251, 62172449), the Natural Science Foundation of Hunan Province (No. 2021JJ40783). This work was supported in part by the High Performance Computing Center of Central South University.

% References should be produced using the bibtex program from suitable
% BiBTeX files (here: strings, refs, manuals). The IEEEbib.bst bibliography
% style file from IEEE produces unsorted bibliography list.
% -------------------------------------------------------------------------
\bibliographystyle{IEEEbib}
\bibliography{reference}

\begin{thebibliography}{10}

\bibitem{HaoranShi2017TowardsAI}
Haoran Shi, Pengtao Xie, Zhiting Hu, Ming Zhang, and Eric~P Xing,
\newblock ``Towards automated icd coding using deep learning,''
\newblock {\em arXiv preprint arXiv:1711.04075}, 2017.

\bibitem{JamesMullenbach2018ExplainablePO}
James Mullenbach, Sarah Wiegreffe, Jon Duke, Jimeng Sun, and Jacob Eisenstein,
\newblock ``Explainable prediction of medical codes from clinical text,''
\newblock in {\em Proceedings of the NAACL 2018: Human Language Technologies,
  Volume 1 (Long Papers)}, June 2018.

\bibitem{ThanhVu2020ALA}
Thanh Vu, Dat~Quoc Nguyen, and Anthony Nguyen,
\newblock ``A label attention model for icd coding from clinical text,''
\newblock in {\em Proceedings of the Twenty-Ninth International Joint
  Conference on Artificial Intelligence, {IJCAI-20}}, Christian Bessiere, Ed.,
  7 2020, pp. 3335--3341.

\bibitem{LeiboLiu2022HierarchicalLA}
Leibo Liu, Oscar Perez-Concha, Anthony Nguyen, Vicki Bennett, and Louisa Jorm,
\newblock ``Hierarchical label-wise attention transformer model for explainable
  icd coding,''
\newblock {\em Journal of biomedical informatics}, vol. 133, pp. 104161, 2022.

\bibitem{JacobDevlin2018BERTPO}
Jacob Devlin, Ming-Wei Chang, Kenton Lee, and Kristina Toutanova,
\newblock ``{BERT}: Pre-training of deep bidirectional transformers for
  language understanding,''
\newblock in {\em Proceedings of the NAACL 2019}, June 2019, pp. 4171--4186.

\bibitem{FeiLi2019ICDCF}
Fei Li and Hong Yu,
\newblock ``Icd coding from clinical text using multi-filter residual
  convolutional neural network,''
\newblock in {\em proceedings of the AAAI}, 2020, vol.~34, pp. 8180--8187.

\bibitem{YangLiu2021EffectiveCA}
Yang Liu, Hua Cheng, Russell Klopfer, Matthew~R. Gormley, and Thomas Schaaf,
\newblock ``Effective convolutional attention network for multi-label clinical
  document classification,''
\newblock in {\em Proceedings of the EMNLP 2021}, Online and Punta Cana,
  Dominican Republic, Nov. 2021, pp. 5941--5953, Association for Computational
  Linguistics.

\bibitem{JieHu2018SqueezeandExcitationN}
Jie Hu, Li~Shen, and Gang Sun,
\newblock ``Squeeze-and-excitation networks,''
\newblock in {\em Proceedings of the CVPR}, 2018, pp. 7132--7141.

\bibitem{SeppHochreiter1997LongSM}
Sepp Hochreiter and J{\"u}rgen Schmidhuber,
\newblock ``Long short-term memory,''
\newblock {\em Neural computation}, vol. 9, no. 8, pp. 1735--1780, 1997.

\bibitem{AlistairEWJohnson2016MIMICIIIAF}
Alistair~EW Johnson, Tom~J Pollard, Lu~Shen, Li-wei~H Lehman, Mengling Feng,
  Mohammad Ghassemi, Benjamin Moody, Peter Szolovits, Leo Anthony~Celi, and
  Roger~G Mark,
\newblock ``Mimic-iii, a freely accessible critical care database,''
\newblock {\em Scientific data}, vol. 3, no. 1, pp. 1--9, 2016.

\bibitem{ben-zaken-etal-2022-bitfit}
Elad Ben~Zaken, Yoav Goldberg, and Shauli Ravfogel,
\newblock ``{B}it{F}it: Simple parameter-efficient fine-tuning for
  transformer-based masked language-models,''
\newblock in {\em ACL}, Dublin, Ireland, May 2022, pp. 1--9, Association for
  Computational Linguistics.

\bibitem{PengfeiCao2020HyperCoreHA}
Pengfei Cao, Yubo Chen, Kang Liu, Jun Zhao, Shengping Liu, and Weifeng Chong,
\newblock ``{H}yper{C}ore: Hyperbolic and co-graph representation for automatic
  {ICD} coding,''
\newblock in {\em Proceedings of the 58th ACL}, Online, July 2020, pp.
  3105--3114, Association for Computational Linguistics.

\bibitem{TongZhou2021AutomaticIC}
Tong Zhou, Pengfei Cao, Yubo Chen, Kang Liu, Jun Zhao, Kun Niu, Weifeng Chong,
  and Shengping Liu,
\newblock ``Automatic {ICD} coding via interactive shared representation
  networks with self-distillation mechanism,''
\newblock in {\em Proceedings of the 59th ACL and the 11th IJCNLP (Volume 1:
  Long Papers)}, Online, Aug. 2021, pp. 5948--5957, Association for
  Computational Linguistics.

\bibitem{ZhilinYang2019XLNetGA}
Zhilin Yang, Zihang Dai, Yiming Yang, Jaime Carbonell, Russ~R Salakhutdinov,
  and Quoc~V Le,
\newblock ``Xlnet: Generalized autoregressive pretraining for language
  understanding,''
\newblock {\em Advances in NeurIPS}, vol. 32, 2019.

\bibitem{XianchengXie2019EHRCW}
Xiancheng Xie, Yun Xiong, Philip~S Yu, and Yangyong Zhu,
\newblock ``Ehr coding with multi-scale feature attention and structured
  knowledge graph propagation,''
\newblock in {\em Proceedings of the 28th CIKM}, 2019, pp. 649--658.

\bibitem{WeiSun2021MultitaskBA}
Wei Sun, Shaoxiong Ji, Erik Cambria, and Pekka Marttinen,
\newblock ``Multitask balanced and recalibrated network for medical code
  prediction,''
\newblock {\em ACM Transactions on Intelligent Systems and Technology}, vol.
  14, no. 1, pp. 1--20, 2022.

\bibitem{ByungHakKim2021ReadAA}
Byung-Hak Kim and Varun Ganapathi,
\newblock ``Read, attend, and code: pushing the limits of medical codes
  prediction from clinical notes by machines,''
\newblock in {\em Machine Learning for Healthcare Conference}. PMLR, 2021, pp.
  196--208.

\bibitem{NingZhang2022HierarchicalBF}
Ning Zhang and Maciej Jankowski,
\newblock ``Hierarchical bert for medical document understanding,''
\newblock {\em arXiv preprint arXiv:2204.09600}, 2022.

\bibitem{MalteFeucht2021DescriptionbasedLA}
Malte Feucht, Zhiliang Wu, Sophia Althammer, and Volker Tresp,
\newblock ``Description-based label attention classifier for explainable
  {ICD}-9 classification,''
\newblock in {\em Proceedings of the W-NUT 2021}, Nov. 2021, pp. 62--66.

\bibitem{ZhengYuan2022CodeSD}
Zheng Yuan, Chuanqi Tan, and Songfang Huang,
\newblock ``Code synonyms do matter: Multiple synonyms matching network for
  automatic {ICD} coding,''
\newblock in {\em Proceedings of the 60th ACL (Volume 2: Short Papers)}, May
  2022, pp. 808--814.

\bibitem{JoonLee2011OpenaccessMD}
Joon Lee, Daniel~J Scott, Mauricio Villarroel, Gari~D Clifford, Mohammed Saeed,
  and Roger~G Mark,
\newblock ``Open-access mimic-ii database for intensive care research,''
\newblock in {\em 2011 Annual International Conference of the IEEE EMBS}. IEEE,
  2011, pp. 8315--8318.

\end{thebibliography}

\end{document}